\documentclass[11pt]{article}

\usepackage{acl}

\usepackage{times}
\usepackage{latexsym}

\usepackage[T1]{fontenc}

\usepackage[utf8]{inputenc}

\usepackage{times}
\usepackage{latexsym}

\usepackage{graphicx}
\usepackage{float}
\usepackage{amsmath}
\usepackage{diagbox}
\usepackage{multirow}
\usepackage{pifont}
\newcommand{\cmark}{\ding{51}}%
\newcommand{\xmark}{\ding{55}}%

\usepackage{microtype}

\title{GPL: Generative Pseudo Labeling for Unsupervised Domain Adaptation of Dense Retrieval}

\author{
Kexin Wang\textsuperscript{\textnormal{1}}, 
Nandan Thakur\textsuperscript{\textnormal{2}}\thanks{~~Contributions made during the stay at the UKP Lab.},
Nils Reimers\textsuperscript{\textnormal{3}},
Iryna Gurevych\textsuperscript{\textnormal{1}}\\
 \textsuperscript{\textnormal{1}} Ubiquitous Knowledge Processing Lab, Technical University of Darmstadt \\
 \textsuperscript{\textnormal{2}} University of Waterloo, \textsuperscript{\textnormal{3}} Hugging Face\\
 \url{www.ukp.tu-darmstadt.de}
}

\begin{document}
\maketitle

\begin{abstract}
Dense retrieval approaches can overcome the lexical gap and lead to significantly improved search results. However, they require large amounts of training data which is not available for most domains. As shown in previous work~\citep{thakur2021beir}, the performance of dense retrievers severely degrades under a domain shift. This limits the usage of dense retrieval approaches to only a few domains with large training datasets.

In this paper, we propose the novel unsupervised domain adaptation method \textit{Generative Pseudo Labeling} (GPL), which combines a query generator with pseudo labeling from a cross-encoder. On six representative domain-specialized datasets, we find the proposed GPL can outperform an out-of-the-box state-of-the-art dense retrieval approach by up to 9.3 points nDCG@10. GPL requires less (unlabeled) data from the target domain and is more robust in its training than previous methods.

We further investigate the role of six recent pre-training methods in the scenario of domain adaptation for retrieval tasks, where only three could yield improved results. The best approach, TSDAE~\cite{wang2021tsdae} can be combined with GPL, yielding another average improvement of 1.4 points nDCG@10 across the six tasks. The code and the models are available
\footnote{\url{https://github.com/UKPLab/gpl}}.

\end{abstract}

\section{Introduction}
Information Retrieval (IR) is a central component of many natural language applications. Traditionally, lexical methods~\citep{DBLP:conf/trec/RobertsonWJHG94} have been used to search through text content. However, these methods suffer from the lexical gap~\citep{DBLP:conf/sigir/BergerCCFM00} and are not able to recognize synonyms and distinguish between ambiguous words. 

Recently, information retrieval methods based on dense vector spaces have become popular to address these challenges. These dense retrieval methods map queries and passages\footnote{We use passage to refer to text of any length.} to a shared, dense vector space and retrieve relevant hits by nearest-neighbor search. Significant improvement over traditional approaches has been shown for various tasks \citep{karpukhin-etal-2020-dense,xiong2021approximate}. This method is also adapted increasingly by industry to enhance the search functionalities of various applications~\citep{choi2020semantic,DBLP:conf/kdd/HuangSSXZPPOY20}.

However, as shown in \citet{thakur2021beir}, dense retrieval methods require large amounts of training data to work well.\footnote{For reference, the popular MS MARCO dataset~\citep{bajaj2018ms} has about 500k training instances; the Natural Questions dataset~\citep{kwiatkowski-etal-2019-natural} has more than 100k training instances. } Most importantly, dense retrieval methods are extremely sensitive to domain shifts: Models trained on MS MARCO perform rather poorly for questions for COVID-19 scientific literature~\citep{wang-etal-2020-cord,10.1145/3451964.3451965}. The MS MARCO dataset was created before COVID-19, hence, it does not include any COVID-19 related topics and models  did not learn how to represent this topic well in a vector space.

\begin{figure*}[t]
  \centering
  \includegraphics[width=140mm]{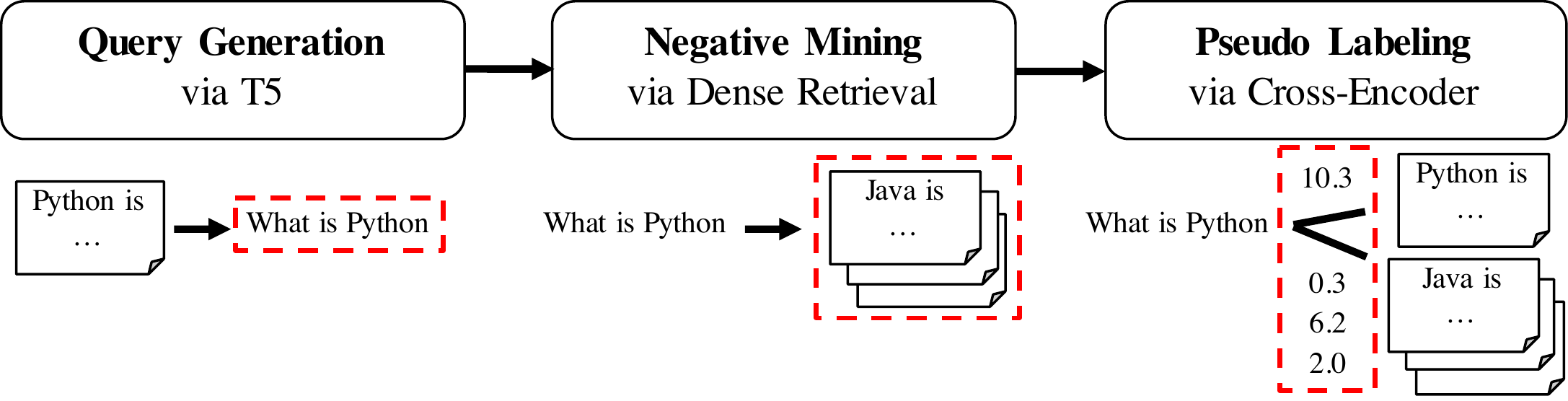}
  \caption{Generative Pseudo Labeling (GPL) for training domain-adapted dense retriever. First, synthetic queries are generated for each passage from the target corpus. Then, the generated queries are used for mining negative passages. Finally, the query-passage pairs are labeled by a cross-encoder and used to train the domain-adapted dense retriever. The output at each step is marked with dashed boxes. }
  \label{fig:gpl}
\end{figure*}

In this work, we present \textit{Generative Pseudo Labeling} (GPL), an unsupervised domain adaptation technique for dense retrieval models (see \autoref{fig:gpl}). For a collection of  paragraphs from the desired domain, we use an existing pre-trained T5 encoder-decoder to generate suitable queries. For each generated query, we retrieve the most similar paragraphs using an existing dense retrieval model which will serve as negative passages. Finally, we use an existing cross-encoder to score each (query, passage)-pair and train a dense retrieval model on these generated, pseudo-labeled queries using MarginMSE-Loss~\cite{DBLP:journals/corr/abs-2010-02666}.

We use publicly available models for query generation, negative mining, and the cross-encoder, which have been trained on the MS MARCO dataset~\cite{bajaj2018ms}, a large-scale dataset from Bing search logs combined with relevant passages from diverse web sources. We evaluate GPL on six representative domain-specific datasets from the BeIR benchmark~\cite{thakur2021beir}. \textbf{GPL improves the performance by up to 9.3 points nDCG@10} compared to state-of-the-art model trained solely on MS MARCO. Compared to the previous state-of-the-art domain-adaption method QGen \cite{ma-etal-2021-zero,thakur2021beir}, GPL improves the performance by up to 4.5 nDCG@10 points. Training with GPL is easy, fast, and data efficient.

We further analyze the role of six recent pre-training methods in the scenario of domain adaptation for retrieval tasks. The best approach is TSDAE \cite{wang2021tsdae}, that outperforms the second best approach, Masked Language Modeling \cite{devlin19BERT} on average by 2.5 points nDCG@10. TSDAE can be combined with GPL, yielding another average improvement of 1.4 point nDCG@10.

\section{Related Work}
\textbf{Pre-Training based Domain Adaptation.} The most common domain adaption technique for transformer models is domain-adaptive pre-training~\cite{gururangan-etal-2020-dont}, which continues pre-training on in-domain data before fine-tuning with labeled data. However, for retrieval it is often difficult to get in-domain labeled data and models are be applied in a zero-shot setting on a given corpus. 
Besides Masked Language Modeling (MLM) \cite{devlin19BERT}, different pre-trained strategies specifically for dense retrieval methods have been proposed. Inverse Cloze Task (ICT)~\cite{lee-etal-2019-latent} generates query-passage pair by randomly selecting one sentence from the passage as the query and the other part as the paired passage. ConDensor (CD)~\cite{DBLP:journals/corr/abs-2104-08253} applies MLM on top of the CLS token embedding from the final layer and the other context embeddings from a previous layer to force the model to learn meaningful CLS representation. SimCSE~\cite{gao-etal-2021-simcse,liu-etal-2021-fast} passes the same input twice through the network with different dropout masks and minimizes the distance of the resulting embeddings, while Contrastive Tension (CT)~\cite{carlsson2021semantic} passes the input through two different models. TSDAE \cite{wang2021tsdae} uses a denoising auto-encoder architecture with bottleneck: Words from the input text are removed and passed through an encoder to generate a fixed-sized embedding. A decoder must reconstruct the original text without noise. As we show in \autoref{sec:unsupervised}, just using these unsupervised techniques is not sufficient and the resulting models perform poorly.

So far, ICT and CD have only been studied on in-domain performance, i.e.\ a large in-domain labeled dataset is available which is used for subsequent supervised fine-tuning. SimCSE, CT, and TSDAE have been only studied for unsupervised sentence embedding learning. As our results show in \autoref{sec:unsupervised}, they do not work at all for purely unsupervised dense retrieval. 

If these pre-training approaches can be used for unsupervised domain adaptation for dense retrieval was so far unclear. In this work, we transfer the setup from \newcite{wang2021tsdae} to dense retrieval and first pre-train on the target corpus, followed by supervised training on labeled data from MS MARCO~\cite{bajaj2018ms}. Performance is then measured on the target corpus.

\textbf{Query Generation.}
Query generation has been used to improve retrieval performances. Doc2query \cite{nogueira2019doc2query,DBLP:journals/corr/abs-1904-08375} expands passages with predicted queries, generated by a trained encoder-decoder model, and uses traditional BM25 lexical search. This performed well in the zero-shot retrieval benchmark BeIR \cite{thakur2021beir}. \citet{ma-etal-2021-zero} proposes QGen, that uses a query generator trained on general domain data to synthesize domain-targeted queries for the target corpus, on which a dense retriever is trained from scratch. As a concurrent work, \citet{embedding-basaed-zero-shot} also proposes the similar method. Following this idea, \citet{thakur2021beir} views QGen as a post-training method to adapt powerful MS MARCO retrievers to the target domains.

Despite the success of QGen, previous methods only consider the cross-entropy loss with in-batch negatives, which provides coarse-grained relevance and thus limits the performance. In this work, we show that extending this approach by using pseudo-labels from a cross-encoder together with hard negatives can boost the performance by several points nDCG@10.

\textbf{Other Methods.} Recently, \citet{xin2021zeroshot} proposes MoDIR to use Domain Adversarial Training (DAT)~\cite{DBLP:journals/jmlr/GaninUAGLLML16} for unsupervised domain adaptation of dense retrievers. MoDIR trains models by generating domain invariant representations to attack a domain classifier. However, as argued in~\citet{karouzos-etal-2021-udalm}, DAT trains models by minimizing the distance between representations from different domains and such learning objective can result in bad embedding space and unstable performance. For sentiment classification, \citet{karouzos-etal-2021-udalm} proposes UDALM based on multiple stages of training. UDALM first applies MLM training on the target domain; and it then applies multi-task learning on the target domain with MLM and on the source domain with a supervised objective. However, as shown in \autoref{sec:results}, we find this method cannot yield improvement for retrieval tasks.

\textbf{Pseudo Labeling and Cross-Encoders:}
Bi-Encoders map queries and passage independently to a shared vector space from which the query-passage similarity is computed. In contrast,  cross-encoders~\cite{Humeau2020Poly-encoders} work on the concatenation of the query and passage and predict a relevance score using cross-attention between query and passage. This can be used in a re-ranking setup \cite{DBLP:journals/corr/abs-1901-04085}, where the relevancy is predicted for all query-passage-pairs for a small candidate set. Previous work has shown that cross-encoders achieve much higher performances  \cite{thakur-etal-2021-augmented,DBLP:journals/corr/abs-2010-02666,ren-etal-2021-rocketqav2} and are less prone to domain shifts \cite{thakur2021beir}. But cross-encoders come with an extremely high computational overhead, making them less suited for a production setting. Transferring knowledge from cross-encoder to bi-encoders have been shown previous for sentence embeddings \cite{thakur-etal-2021-augmented} and for dense retrieval: \citet{DBLP:journals/corr/abs-2010-02666} predict cross-encoder scores for (query, positive)-pairs and (query, negative)-pairs and learns a bi-encoder to predict the margin between the two scores. This has been shown highly effective for in-domain dense retrieval.

\section{Method}
This section describes our proposed \textit{Generative Pseudo Labeling} (GPL) method for the unsupervised domain adaptation of dense retrievers. \autoref{fig:gpl} illustrates the idea of GPL. 

For a given target corpus, we generate for each passage three queries (cf.\ \autoref{tbl:how_many_queries}) using an T5-encoder-decoder model~\cite{JMLR:v21:20-074}. For each of the generated queries, we use an existing retrieval system to retrieve 50 negative passages. Dense retrieval with a pre-existing model was slightly more effective than BM25 lexical retrieval (cf.\ \autoref{sec:switching_negative_mining_retrievers}). For each (query, positive, negative)-tuple we compute the margin $\delta = \mathrm{CE}(Q, P^+,) - \mathrm{CE}(Q, P^-)$ with $\mathrm{CE}$ the score as predicted by a cross-encoder, $Q$ the query and $P^+ / P^-$  the positive / negative passage.

We use the synthetic dataset $D_{\mathrm{GPL}}=\{(Q_i, P_i, P^-_i, \delta_i)\}_{i}$ with the MarginMSE loss \cite{DBLP:journals/corr/abs-2010-02666} for training a domain-adapted dense retriever that maps queries and passages into the shared vector space.

Our method requires from the target domain just an unlabeled collection of passages. Further, we use use pre-existing T5- and cross-encoder models that have been trained on the MS MARCO passages dataset.

\label{sec:background}
\textbf{Query Generation:} To enable supervised training on the target corpus, synthetic queries can be generated for the target passages using a query generator trained on a different, existing dataset like MS MARCO. Previous work QGen~\cite{ma-etal-2021-zero} used the simple MultipleNegativesRanking (MNRL) loss~\cite{DBLP:journals/corr/HendersonASSLGK17,DBLP:journals/corr/abs-1807-03748} with in-batch negatives to train the model:

\begin{align*}
L_{\mathrm{MNRL}}&(\theta) = \\ -\frac{1}{M}\sum_{i=0}^{M-1}&\log\frac{\exp\big(\tau\cdot {\sigma(f_{\theta}(Q_i), f_{\theta}(P_i))}\big)}{\sum_{j=0}^{M-1}\exp\big(\tau\cdot \sigma(f_{\theta}(Q_i), f_{\theta}(P_j))\big)}
\end{align*}
where $P_i$ is a relevant passage for $Q_i$; $\sigma$ is a certain similarity function for vectors; $\tau$ controls the sharpness of the softmax normalization; $M$ is the batch size.

\textbf{MarginMSE loss:} MultipleNegativesRanking loss considers only the coarse relationship between queries and passages, i.e.\ the matching passage is considered as relevant while all other passages are considered irrelevant. However, the query encoder is not without flaws and might generate queries that are not answerable by the passage. Further, other passages might actually be relevant as well for a given query, which is especially the case if training is done with hard negatives as we do it for GPL. 

In contrast, MarginMSE loss~\cite{DBLP:journals/corr/abs-2010-02666} uses a powerful cross-encoder to soft-label (query, passage) pairs. It then teaches the dense retriever to mimic the score margin between the positive and negative query-passage pairs. Formally, 
\begin{equation}
\label{eq:margin-mse}
L_{\mathrm{MarginMSE}}(\theta) = -\frac{1}{M}\sum_{i=0}^{M-1}|\hat{\delta_i}- \delta_i|^2
\end{equation}where $\hat{\delta_i}$ is the corresponding score margin of the student dense retriever, i.e. $\hat{\delta_i}=f_{\theta}(Q_i)^T f_{\theta}(P_i)-f_{\theta}(Q_i)^T f_{\theta}(P_i^-)$. Here the dot-product is usually used due to the infinite range of the cross-encoder scores. 

This loss is a critical component of GPL, as it solves two major issues from the previous QGen method: A badly generated query for a given passage will get a low score from the cross-encoder, hence, we do not expect the dense retriever to put the query and passage close in the vector space. A false negative will lead to a high score from the cross-encoder, hence, we do not force the dense retriever to assign a large distance between the corresponding embeddings. In section~\ref{sec:robust_query_generation}, we show that GPL is a lot more robust to badly generated queries than the previous QGen method.

\section{Experiments}
\label{sec:exps}
In this section, we describe the experimental setup, the datasets used and the baselines for comparison.

\subsection{Experimental Setup}
We use the MS MARCO passage ranking dataset~\cite{bajaj2018ms} as the data from the source domain. It has 8.8M passages and 532.8K query-passage pairs labeled as relevant in the training set. As \autoref{tbl:main_results} shows, a state-of-the-art dense retrieval model, achieving an MRR@10 of 33.2 points on the MS MARCO passage ranking dataset, performs poorly on the six selected domain-specific retrieval datasets when compared to simple BM25 lexical search.

We use the DistilBERT~\cite{DBLP:journals/corr/abs-1910-01108} for all the experiments. We use the concatenation of the title and the body text as the input passage for all the models. We use a maximum sequence length of 350 with mean pooling and dot-product similarity by default. For QGen, we use the default setting in \citet{thakur2021beir}: 1-epoch training and batch size 75. For GPL, we train the models with 140k training steps and batch size 32. To generate queries for both QGen and GPL, we use the DocT5Query~\cite{nogueira2019doc2query} generator trained on MS MARCO and generate~\footnote{We use the script from BeIR at \url{https://github.com/UKPLab/beir}.} queries using nucleus sampling with temperature 1.0, $k=25$ and $p=0.95$. To retrieve hard negatives for both GPL and the zero-shot setting of MS MARCO training, we use two dense retrievers with cosine-similarity trained on MS MARCO: \textit{msmarco-distilbert-base-v3} and \textit{msmarco-MiniLM-L-6-v3} from Sentence-Transformers\footnote{\url{https://github.com/UKPLab/sentence-transformers}}. The zero-shot performance of these two dense retrievers are available in \autoref{sec:retrievers_in_gpl}. We retrieve 50 negatives using each retriever and uniformly sample one negative passage and one positive passage for each training query to form one training example. For pseudo labeling, we use the \textit{ms-marco-MiniLM-L-6-v2}\footnote{\url{https://huggingface.co/cross-encoder/ms-marco-MiniLM-L-6-v2}} cross-encoder. For all the pre-training methods (e.g. TSDAE and MLM), we train the models for 100K training steps and with batch size 8. 

As shown in Section~\ref{sec:analysis}, small corpora require more generated queries and for large corpora, a small down-sampled subset (e.g. 50K) is enough for good performance. Based on these findings, we adjust the number of generated queries per passage $q_{\mathrm{avg.}}$ and the corpus size $|C|$ to make the total number of generated queries equal to a fixed number, 250K, i.e. $q_{\mathrm{avg.}}\times|C|=250\mathrm{K}$. In detail, we first set $q_{\mathrm{avg.}} >= 3$ and uniformly down-sample the corpus if $3\times |C|>250\mathrm{K}$; then we calculate $q_{\mathrm{avg.}}=\lceil250\mathrm{K}/|C|\rceil$.
For example, the $q_{\mathrm{avg.}}$ values for FiQA (original size = 57.6K) and Robust04 (original size = 528.2K) are 5 and 3, resp. and the Robust04 corpus is down-sampled to 83.3K. QGen and GPL share the generated queries for fair comparision.

\subsection{Evaluation}
As our methods focus on domain adaptation to specialized domains, we selected six domain-specific text retrieval tasks from the BeIR benchmark~\cite{thakur2021beir}: FiQA (financial domain) \cite{fiqa-2018}, SciFact (scientific papers) \cite{scifact-2020}, BioASQ (biomedical Q\&A) \cite{bioasq-2015}, TREC-COVID (scientific papers on COVID-19) \cite{trec-covid-2020}, CQADupStack (12 StackExchange subforums) \cite{cqadupstack-2015} and Robust04 (news articles)  \cite{robust04-2005}. These selected datasets each contain a corpus with a rather specific language and can thus act as a suitable test bed for domain adaptation. 

The detailed information for all the target datasets is available at \autoref{sec:target_datasets}. We make modification on BioASQ and TREC-COVID. For efficient training and evaluation on BioASQ, we randomly remove irrelevant passages to make the final corpus size to 1M. In TREC-COVID, the original corpus has many documents with a missing abstract. The retrieval systems that were used to create the annotation pool for TREC-COVID often ignored such documents, leading to a strong annotation bias for these documents. Hence, we removed all documents with a missing abstract from the corpus. The evaluation results on the original BioASQ and TREC-COVID are available at~\autoref{sec:full_beir}. Evaluation is done using nDCG@10.

\begin{table*}[t]
\centering
\resizebox{14cm}{!}{
\begin{tabular}{|l|c|c|c|c|c|c|c|} 
\hline
\diagbox{\textbf{Method}}{\textbf{Dataset}} & \textbf{FiQA}          & \textbf{SciFact}       & \textbf{BioASQ} & \textbf{TRECC.}        & \textbf{CQADup.}       & \textbf{Robust04}      & \textbf{Avg.}        \\ 
\hline
\multicolumn{8}{|l|}{\textit{Zero-Shot Models}}                                                                                                                                                                      \\ 
\hline
MS MARCO                & 26.7 & 57.1 & 52.9 & 66.1 & 29.6 & 39.0 & 45.2 \\
PAQ         & 15.2 & 53.3 & 44.0 & 23.8 & 24.5 & 31.9 & 32.1 \\
PAQ + MS MARCO                         & 26.7 & 57.6 & 53.8 & 63.4 & 30.6 & 37.2 & 44.9 \\
{TSDAE}$_{\text{MS MARCO}}$                         & 26.7 & 55.5 & 51.4 & 65.6 & 30.5 & 36.6 & 44.4 \\ 
BM25                                      & 23.9                   & 66.1                   & 70.7            & 60.1                   & 31.5                   & 38.7                   & 48.5   \\
\hline
\multicolumn{8}{|l|}{\textit{Previous Domain Adaptation Methods}}                                                                                                                                                                       \\ 
\hline

UDALM                                      & 23.3 & 33.6 & 33.1 & 57.1 & 24.6 & 26.3 & 33.0 \\
MoDIR                                     & 29.6 & 50.2 & 47.9 & 66.0 & 29.7 & -- & -- \\
\hline
\multicolumn{8}{|l|}{\textit{Pre-Training based Domain Adaptation: Target $\to$ MS MARCO}}                                                                                                                           \\ 
\hline
CT & 28.3 & 55.6 & 49.9 & 63.8 & 30.5 & 35.9 & 44.0 \\
CD & 27.0 & 62.7 & 47.7 & 65.4 & 30.6 & 34.5 & 44.7 \\
SimCSE & 26.7 & 55.0 & 53.2 & 68.3 & 29.0 & 37.9 & 45.0 \\
ICT & 27.0 & 58.3 & 55.3 & 69.7 & 31.3 & 37.4 & 46.5 \\
MLM & 30.2 & 60.0 & 51.3 & 69.5 & 30.4 & 38.8 & 46.7 \\
TSDAE & 29.3 & 62.8 & 55.5 & \textbf{76.1} & 31.8 & 39.4 & 49.2 \\
\hline
\multicolumn{8}{|l|}{\textit{Generation-based Domain Adaptation (Previous State-of-the-Art)}}                                                                                                                        \\ 
\hline
QGen    & 28.7 & 63.8 & 56.5 & 72.4 & 33.0 & 38.1 & 48.8 \\
QGen (w/ Hard Negatives) & 26.0 & 59.6 & 57.7 & 65.0 & 33.2 & 36.5 & 46.3 \\
TSDAE + QGen (Ours) & 31.4 & 66.7 & 58.1 & 72.6 & \textbf{35.3} & 37.4 & 50.3 \\
\hline
\multicolumn{8}{|l|}{\textit{Proposed Method: Generative Pseudo Labeling}}                                                                                                                                           \\ 
\hline
GPL & 32.8 & 66.4 & 61.0 & 72.6 & 34.5 & 41.4 & 51.5 \\
TSDAE + GPL &  \textbf{34.4} & \textbf{68.9} & \textbf{61.6} & 74.6 & 35.1 & \textbf{43.0} & \textbf{52.9} \\
\hline
\multicolumn{8}{|l|}{\textit{Re-Ranking with Cross-Encoders (Upper Bound, Inefficient at Inference)}}                                                                                                                                                        \\ 
\hline
BM25 + CE   & 33.1 & 67.6 & 72.8 & 71.2 & 36.8 & 46.7 & 54.7 \\
MS MARCO + CE   & 33.0 & 66.9 & 57.4 & 65.1 & 36.9 & 44.7 & 50.7 \\
TSDAE + GPL + CE    & 36.4 & 68.3 & 68.0 & 71.4 & 38.1 & 48.3 & 55.1 \\
\hline
\end{tabular}}
\caption{Evaluation using nDCG@10. The best results of the single-stage dense retrievers are bold. TRECC. and CQADup. are short for TREC-COVID and CQADupStack. Our proposed GPL significantly outperforms other domain adaptation methods. For the first time, we investigate the TSDAE pre-training in domain adaptation for dense retrieval and find it can significantly improve both QGen and GPL. The results on the full 18 BeIR datasets can be found at~\autoref{sec:full_beir}.}
\label{tbl:main_results}
\end{table*}

\subsection{Baselines}

\textbf{Zero-Shot Models:} We apply supervised training on MS MARCO or PAQ~\cite{DBLP:journals/corr/abs-2102-07033} and evaluate the trained retrievers on the target datasets. (a) \textbf{MS MARCO} represents a distilbert-base dense retrieval model trained with MarginMSE on the MS MARCO dataset with batch-size 75 for 70k steps. (b) \textbf{PAQ}~\cite{DBLP:journals/corr/abs-2107-13602} represents MNRL training on the PAQ dataset. (c) \textbf{PAQ + MS MARCO} represents MNRL training on PAQ followed by MarginMSE training on MS MARCO. (d) \textbf{TSDAE}$_{\textbf{MS MARCO}}$ represents TSDAE~\cite{wang2021tsdae} pre-training on MS MARCO followed by MarginMSE training on MS MARCO. (e) \textbf{BM25} system based on lexical matching from Elasticsearch\footnote{\url{https://www.elastic.co}}.

\textbf{Previous Domain Adaptation Methods:} We include two previous unsupervised domain adaptation methods, UDALM~\cite{karouzos-etal-2021-udalm} and MoDIR~\cite{xin2021zeroshot}. For \textbf{UDALM}, we apply MLM training on the target corpus and then apply the multi-task training of MarginMSE training on MS MARCO and MLM training on the target corpus. For \textbf{MoDIR}, it starts from the ANCE checkpoint and apply domain adversarial training on MS MARCO and the target dataset. As of writing, the training code of MoDIR is not public, but domain adapted models for 5 out of 6 datasets have been released by the authors.

\textbf{Pre-Training based Domain Adaptation:} We follow the setup proposed in \citet{wang2021tsdae} on domain-adapted pre-training: We pre-train the dense retrievers with different methods on the target corpus and then continue to train the models on MS MARCO with MarginMSE loss. The pre-training methods consist of: (a) \textbf{CD}~\cite{DBLP:journals/corr/abs-2104-08253} extracts the hidden representations from an intermediate layer and applies MLM on the CLS token representation and these extracted hidden representations\footnote{CD can only be applied with CLS pooling.}. (b) \textbf{SimCSE}~\cite{DBLP:journals/corr/abs-2104-08821,liu-etal-2021-fast} simply encode the same text twice with different dropout masks in combination with MNRL loss. (c) \textbf{CT}~\cite{carlsson2021semantic} is similar to SimCSE but it uses two independent encoders to encode a pair of text. (d) \textbf{MLM}~\cite{devlin19BERT} uses the default setting in original paper, where 15\% tokens in a text are sampled to be masked and are needed to be predicted. (e) \textbf{ICT}~\cite{lee-etal-2019-latent} uniformly samples one sentence from a passage as the pseudo query to that passage and uses MNRL loss on the synthetic data. We follow the setting in \citet{lee-etal-2019-latent} and masked out the selected sentence 90\%  of the time. (f) \textbf{TSDAE}~\cite{wang2021tsdae}  uses a denoising autoencoder  to pre-train the dense retrievers with 60\% random tokens deleted in the input texts.

\textbf{Generation-based Domain Adaptation:} We use the training script\footnote{\url{https://github.com/UKPLab/beir}} from \citet{thakur2021beir} to train QGen models with the default setting. Cosine similarity is used and the models are fine-tuned for 1 epoch with MNRL. The default QGen is trained with in-batch negatives. For a fair comparison, we also test QGen with hard negatives as used in GPL, noted as \textbf{QGen (w/ Hard Negatives)}. Further, We we test the combination of TSDAE and QGen (\textbf{TSDAE + QGen}).

\textbf{Re-Ranking with Cross-Encoders:} We also include results of the powerful but inefficient re-ranking methods for reference. Three retrievers for the first-phrase retrieval are tested: BM25 from Elasticsearch, the zero-shot MS MARCO retriever and the enhanced GPL retriever by TSDAE pre-training. We use the cross-encoder \textit{ms-marco-MiniLM-L-6-v2} from Sentence-Transformers, which is also for pseudo labeling for GPL.

\section{Results}
\label{sec:results}

\textbf{Pre-Training based Domain Adaptation:}

The results are shown in \autoref{tbl:main_results}. Compared with the zero-shot MS MARCO model, TSDAE, MLM and ICT can improve the performance if we first pre-train on the target corpus and then perform supervised training on MS MARCO. Among them, TSDAE is the most effective method, outperforming the zero-shot baseline by 
by 4.0 points nDCG@10 on average. CD, CT and SimCSE are not able to adapt to the domains in a pre-training setup and achieve a performance worse than the zero-shot model.

To ensure that TSDAE actually learns domain specific terminology, we include $\text{TSDAE}_{\text{MS MARCO}}$ in our experiments: Here, we performed TSDAE pre-training on the MS MARCO dataset follow by supervised learning on MS MARCO. This performs slightly weaker than the zero-shot MS MARCO model. 

We also tested the pre-training methods without any supervised training on MS MARCO. We find all of them fail miserably compared as shown in \autoref{sec:unsupervised} .

\textbf{Previous Domain Adaptation Methods:} We test MoDIR on the datasets except Robust04\footnote{The original author did not train the model on Robust04 and the code is also not available.}. MoDIR performs on-par with our zero-shot MS MARCO model on FiQA, TREC-COVID and CQADupStack, while it performs much weaker on SciFact and BioASQ. An improved training setup with MoDIR could improve the results.

We also test UDALM, which first does MLM pre-training on the target corpus, and then runs multitask learning with MLM objective and supervised training on MS MARCO. The results show that UDALM in this case greatly harms the performance by 12.2 points in average, when compared with the MLM-pre-training approach. We suppose this is because unlike text classification, the dense retrieval models usually do not have an additional task head and the direct MLM training conflicts with the supervised training.

\textbf{Generation-based Domain Adaptation:} The results show that the previous best method, QGen, can successfully adapt the MS MARCO models to the new domains, improving the performance on average by 3.6 points. It performs on par with TSDAE-based domain-adaptive pre-training. Combining TSDAE with QGen can further improve the performance by 1.5 points. 

When using QGen with hard negatives instead of random in-batch negatives, the performance decreases by 2.5 points in average. QGen is sensitive to false negatives, i.e.\ negative passage that are actually relevant for the query. This is a common issue for hard negative mining. GPL solves this issue by using the cross-encoder to determine the distance between the query and a passage. We give more analysis in \autoref{sec:case_study}.

\textbf{Generative Pseudo Labeling (GPL, proposed method):} We find GPL is significantly better on almost all the datasets compared to the other tested method, outperforming QGen by up to 4.5 points (on BioASQ) and in average by 2.7 points. One exception is TREC-COVID, but as this dataset has just 50 test queries, so this difference can be due to noise. 

As a further enhancement, we find that TSDAE-based domain-adaptive pre-training combined with GPL (i.e.\ TSDAE + GPL) can further improve the performance on all the datasets, achieving the new state-of-the-art result of 52.9 nDCG@10 points in average. It outperforms the out-of-the-box MS MARCO model 7.7 points on average.

For the results of GPL on the full 18 BeIR datasets, please refer to~\autoref{sec:full_beir}.

\textbf{Re-ranking with Cross-Encoders:} Cross-encoders perform well in a zero-shot setting and outperform dense retrieval approaches significantly \cite{thakur2021beir}, but they come with a significant computational cost at inference. TSDAE and GPL can narrow but not fully close the performance gap. Due to the much lower computational costs at inference, the TSDAE + GPL model would be preferable in a production setting.

\section{Analysis}
\label{sec:analysis}
In this section, we analyze the influence of training steps, corpus size, query generation and choices of starting checkpoints on GPL.

\subsection{Influence of Training Steps}

\begin{figure}[t]
  \centering
  \includegraphics[width=60mm]{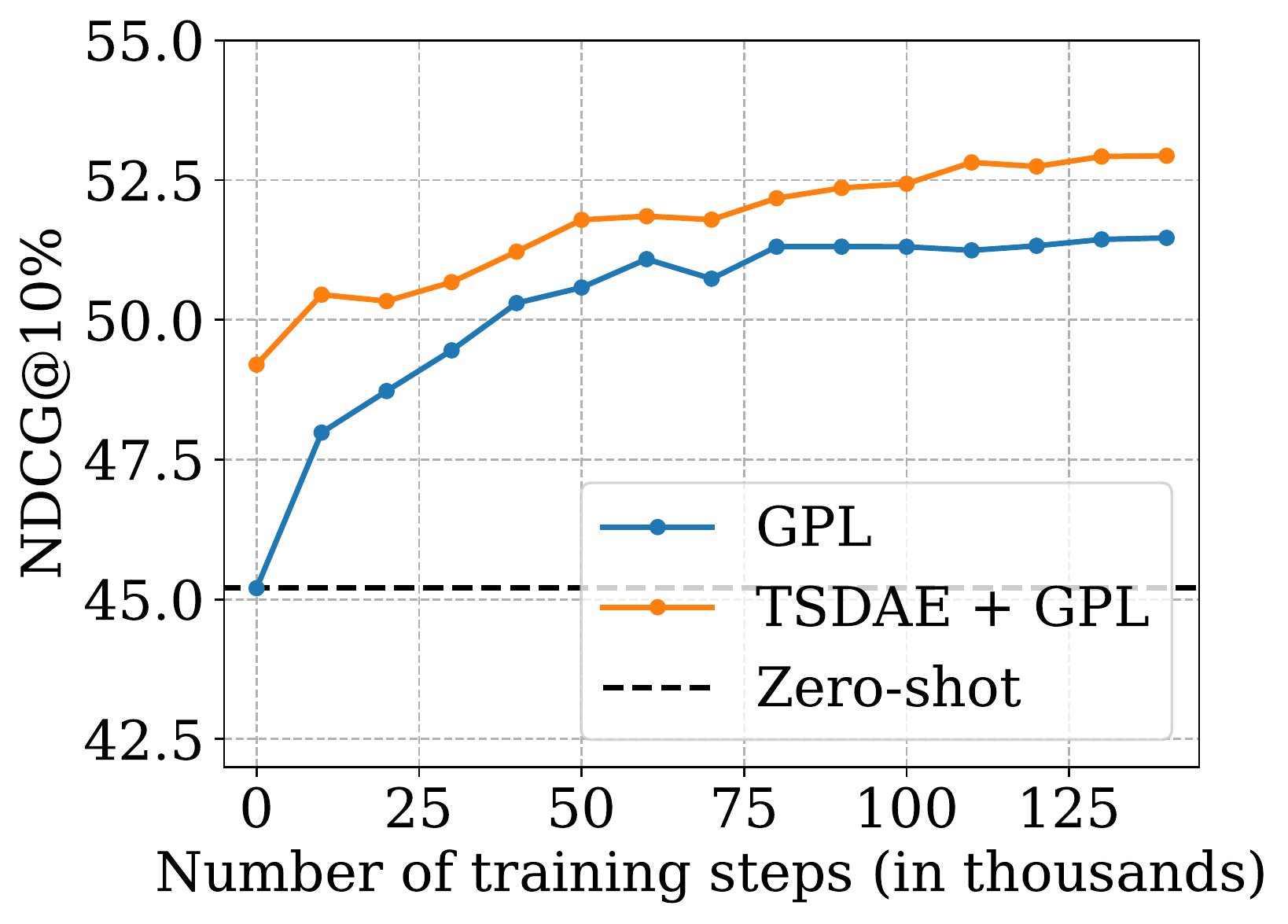}
  \caption{Influence of the number training steps on the averaged performance. The performance of GPL begins to be saturated after 100K steps. TSDAE helps improve the performance during the whole training stage.}
  \label{fig:dynamics}
\end{figure}

We first analyze the influence of the number of training steps on the model performance. We evaluate the models every 10K training steps and end the training after 140K steps. The results for the change of averaged performance on all the datasets are shown in \autoref{fig:dynamics}. We find the performance of GPL begins to be saturated after around 100K steps. With the TSDAE pre-training, the performance can be improved consistently during the whole training stage. For reference, training a distilbert-base model for 100k steps takes about 9.6 hours on a single V100 GPU.

\subsection{Influence of Corpus Size}
\begin{table}[t]
    \centering
    \resizebox{7.5cm}{!}{
    \begin{tabular}{|l|c|c|c|c|c|} 
    \hline
    \diagbox{\textbf{Method}}{\textbf{Size}} & \textbf{1K} & \textbf{10K} & \textbf{50K} & \textbf{250K} & \textbf{528K}  \\ 
    \hline
    QGen      & 35.5 & 36.5 & 38.7 & 37.5 & 37.0 \\
    \hline
    GPL       & 37.6 & 41.4 & 42.5 & 41.4 & 41.3 \\
    \hline
    Zero-shot & \multicolumn{5}{c|}{39.0}                                                  \\
    \hline
    \end{tabular}}
    \caption{Influence of corpus size on performance on Robust04. The full size is 528K. GPL can achieve the best performance with as little as 50K passages.}
    \label{tbl:how_many_documents}
\end{table}

We next analyze the influence of different corpus sizes. We use Robust04 for this analysis, since it has a relatively large size. We sample 1K, 10K, 50K and 250K passages from the whole corpus independently to form small corpora and train QGen and GPL on the same small corpus. The results are shown in \autoref{tbl:how_many_documents}. We find with more than 10K passages, GPL can already significantly outperform the zero-shot baseline by 2.4 NDCG@10 points; with more than 50K passages, the performance begins to saturate. On the other hand, QGen falls behind the zero-shot baseline for each corpus size.

\subsection{Robustness against Query Generation}
\label{sec:robust_query_generation}
\begin{table}[t]
    \centering
    \resizebox{7.5cm}{!}{
    \begin{tabular}{|l|l|l|l|l|l|l|l|l|} 
    \hline
    \multirow{2}{*}{\textbf{Dataset}}                                           & \multirow{2}{*}{\textbf{Method}} & \multicolumn{7}{c|}{\textbf{Queries Per Passage}}                                                                                                                                                                                                    \\ 
    \cline{3-9}
                                                                                &                                  & \multicolumn{1}{c|}{\textbf{1 }} & \multicolumn{1}{c|}{\textbf{2 }} & \multicolumn{1}{c|}{\textbf{3 }} & \multicolumn{1}{c|}{\textbf{5 }} & \multicolumn{1}{c|}{\textbf{10 }} & \multicolumn{1}{c|}{\textbf{25}} & \multicolumn{1}{c|}{\textbf{50}}  \\ 
    \hline
    \multirow{3}{*}{\begin{tabular}[c]{@{}l@{}}SciFact\\(5.2K)\end{tabular}}   & QGen                             & \multicolumn{1}{c|}{56.7}        & \multicolumn{1}{c|}{59.6}        & \multicolumn{1}{c|}{60.2}        & \multicolumn{1}{c|}{59.9}        & \multicolumn{1}{c|}{61.5}         & 62.2                             & 63.7                              \\ 
    \cline{2-9}
                                                                                & GPL                              & \multicolumn{1}{c|}{61.7}        & \multicolumn{1}{c|}{63.2}        & \multicolumn{1}{c|}{63.8}        & \multicolumn{1}{c|}{64.7}        & \multicolumn{1}{c|}{66.8}         & 66.9                             & 67.9                              \\ 
    \cline{2-9}
                                                                                & Zero-shot                        & \multicolumn{7}{c|}{57.1}                                                                                                                                                                                                                            \\ 
    \hline
    \multirow{3}{*}{\begin{tabular}[c]{@{}l@{}}FiQA\\(57.6K)\end{tabular}}       & QGen                             & 27.3                             & 28.1                             & 27.8                             & 28.5                             & 29.3                              & 31.1                             & 31.8                              \\ 
    \cline{2-9}
                                                                                & GPL                              & 31.5                             & 32.2                             & 32.3                             & 32.8                             & 33.0                              & 33.5                             & 33.5                              \\ 
    \cline{2-9}
                                                                                & Zero-shot                        & \multicolumn{7}{c|}{26.7}                                                                                                                                                                                                                            \\ 
    \hline
    \multirow{3}{*}{\begin{tabular}[c]{@{}l@{}}Robust04\\(528.2K)\end{tabular}} & QGen                             & 37.9                             & 38.7                             & 37.0                             & 37.3                             & 38.2                              & 37.7                             &     37.7                              \\ 
    \cline{2-9}
                                                                                & GPL                              & 42.0                             & 41.3                             & 41.4                             & 41.2                             & 40.9                              & 41.2                             & 40.6                              \\ 
    \cline{2-9}
                                                                                & Zero-shot                        & \multicolumn{7}{c|}{39.0}                                                                                                                                                                                                                            \\
    \hline
    \end{tabular}
}
    \caption{Influence of number of generated Queries Per Passage (QPP) on the performance on SciFact, FiQA and Robust04. Corpus size is labeled under each dataset name. Smaller corpora, e.g. SciFact and FiQA require larger QPP to achieve the optimal performance.}
    \label{tbl:how_many_queries}
\end{table}

Next, we study how the query generation influences the model performance. First, we train QGen and GPL on SciFact, FiQA and Robust04, with 1 up to 50 generated Queries Per Passage (QPP). The results are shown in \autoref{tbl:how_many_queries}. We observe that smaller corpora, e.g. SciFact (size = 5.2K) and FiQA (size = 57.6K) require more generated queries per passage than the large one, Robust04 (size = 528.2K). For example, GPL needs QPP equal to around 50, 5 and 1 for SciFact, FiQA and Robust04, resp. to achieve the optimal performance.

\begin{figure}[t]
  \centering
  \includegraphics[width=60mm]{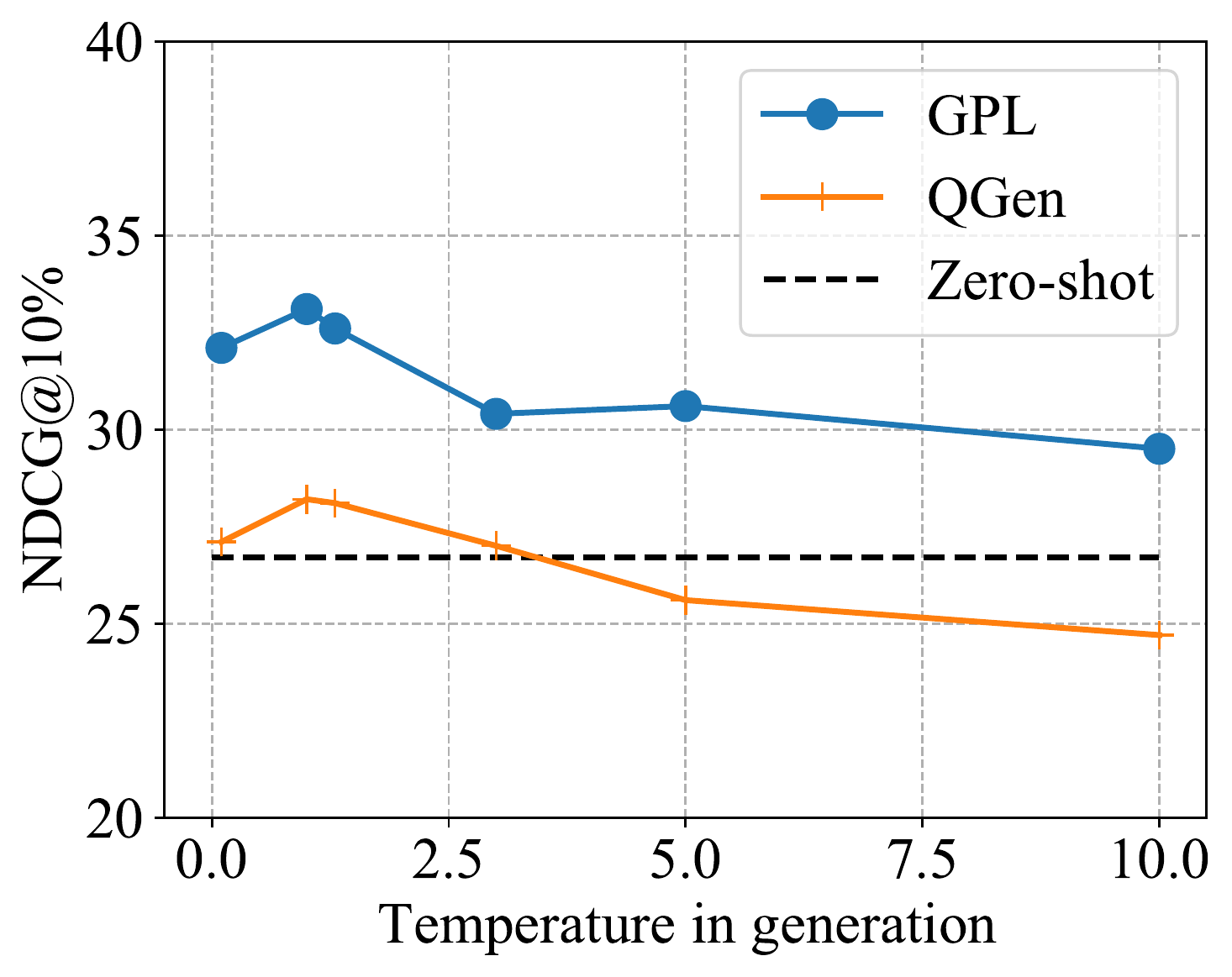}
  \caption{Influence of the temperature in generation on the performance on FiQA. A higher temperature means more diverse queries but of lower quality. GPL can still yield around 3.0-point improvement over the zero-shot baseline with high temperature value of 10.0, where the generated queries have nearly no connection to the passages.}
  \label{fig:temp}
\end{figure}

The temperature plays an important role in nucleus sampling, higher values make the generated queries more diverse, but of lower quality. We train QGen and GPL on FiQA with different temperatures: 0.1, 1, 1.3, 3, 5 and 10. Examples of generated queries are available in~\autoref{sec:temp_examples}. We generated 3 queries per passage. The results are shown in \autoref{fig:temp}. We find the performance of QGen and GPL both peaks at 1.0. With a higher temperature, the next-token distribution will be flatter and more diverse queries, but of lower quality, will be generated. With high temperatures, the generated queries have nearly no relationship to the passage. QGen will perform poorly in these cases, worse than the zero-shot model. In contrast, GPL performs still well even when the generates queries are of such low quality.

\subsection{Sensitivity to Starting Checkpoints}
We also analyze the influence of initialization on GPL. In the default setting, we start from a distilbert-model supervised on MS MARCO using MarginMSE loss. We also evaluate to directly fine-tune a distilbert-model using QGen, GPL and TSDAE + GPL. The performance averaged on all the datasets are shown in \autoref{tbl:checkpoint}. We find the MS MARCO training has relatively small effect on the performance of GPL (with 0.3-point difference in average), while QGen highly relies on the choice of the initialization checkpoint (with 1.9-point difference in average). 

\begin{table}[t]
\centering
\resizebox{6.5cm}{!}{
\begin{tabular}{|l|c|c|} 
\hline
\diagbox{\textbf{Method}}{\textbf{Init.}} & \textbf{Distilbert} & \textbf{MS MARCO}  \\ 
\hline
QGen                                      & 46.9                & 48.8                 \\ 
\hline
TSDAE + QGen (Ours)                                      & 49.6                & 50.3                 \\ 
\hline
GPL                                       & 51.2                & 51.5                 \\ 
\hline
TSDAE + GPL                                & 52.3                & 52.9                 \\ 
\hline
Zero-shot                                 & --                & 45.2                   \\
\hline
\end{tabular}}
\caption{Influence of initialization checkpoint on performance in average. GPL yields similar performance when starting from different checkpoints.}
\label{tbl:checkpoint}
\end{table}

\section{Case Study: Fine-Grained Labels}
\label{sec:case_study}
GPL uses continuous pseudo labels from a cross-encoder, which can provide more fine-grained information and is more informative than the simple 0-1 labels as in QGen. In this section, we give a more detailed insight into it by a case study.

One example from FiQA is shown in \autoref{tbl:case_study}. The generated query for the positive passage asks for the definition of ``futures contract''. Negative 1 and 2 only mention futures contract without explaining the term (with low GPL labels below 2.0), while Negative 3 gives the required definition (with high GPL label 8.2). As an interesting case, Negative 4 gives a partial explanation of the term (with medium GPL label 6.9). GPL assigns suitable fine-grained labels to different negative passages. In contrast, QGen simply labels all of them as 0, i.e.\ as irrelevant. Such difference explains the advantage of GPL over QGen and why using hard negatives harms the performance of QGen in \autoref{tbl:main_results}.

\begin{table}[t]
\centering
\resizebox{7.6cm}{!}{
\begin{tabular}{|l|c|c|c|} 
\hline
\textbf{Item}  & \textbf{Text}                                                                                                                                                                                        & \textbf{GPL} & \textbf{QGen}  \\ 
\hline
\textbf{Query} & what is \textbf{futures contract}                                                                                                                                                                    & --           & --             \\ 
\hline
\textbf{Positive}     & \begin{tabular}[c]{@{}c@{}}\textbf{Futures contracts} are a \\member of a larger class \\of financial assets called \\derivatives ...\end{tabular}                                                   & 10.3        & 1              \\ 
\hline
\textbf{Negative 1}    & \begin{tabular}[c]{@{}c@{}}... Anyway in this one example \\the s\&p 500 \textbf{futures~contract} \\has an "initial margin" of \\\$19,250,~ meaning ...\end{tabular}                                & 2.0        & 0              \\ 
\hline
\textbf{Negative 2}    & \begin{tabular}[c]{@{}c@{}}... but the moment you exercise \\you must have \$5,940 in a \\margin account to actually \\use the \textbf{futures contract} ...\end{tabular}                            & 0.3         & 0              \\ 
\hline
\textbf{Negative 3}    & \begin{tabular}[c]{@{}c@{}}... a \textbf{futures contract} is simply \\a contract that requires party A\\~to buy a given amount of a \\commodity from party B at~a\\~specified price...\end{tabular} & 8.2          & 0              \\ 
\hline
\textbf{Negative 4}    & \begin{tabular}[c]{@{}c@{}}... A \textbf{futures contract }commits \\two parties to a buy/sell of the \\underlying securities, but ...\end{tabular}                                                  & 6.9         & 0              \\ 
\hline
\end{tabular}}
\caption{Examples of the labels assigned to different query-passage pairs in FiQA by GPL and QGen. The key term "futures contract" are marked in bold. QGen uses only 0-1 scores. GPL uses raw logits, which can be any value between positive and negative infinity.}
\label{tbl:case_study}
\end{table}

\section{Conclusion}
In this work we propose GPL, a novel unsupervised domain adaptation method for dense retrieval models. It generates queries for a target corpus and pseudo labels these with a cross-encoders. Pseudo-labeling overcomes two important short-comings of previous methods: Not all generated queries are of high quality and pseudo-labels efficiently detects those. Further, training with mined hard negatives is possible as the pseudo labels performs efficient denoising.   

We observe GPL performs well on all the datasets and significantly outperforms other approaches. As a limitation, GPL requires a relatively complex training setup and future work can focus on simplify this training pipeline.

In this work, we also evaluated different pre-training strategies in a domain-adaptive pre-training setup: We first pre-trained on the target domain, then performed supervised training on MS MARCO. ICT and MLM were able to yield a small improvement (by <=1.5 nDCG@10 points on average), while TSDAE was able to yield a significant improvement of 4 nDCG@10 points on average. Other approaches degraded the performance.

\section*{Acknowledgments}
This work has been funded by the German Research Foundation (DFG) as part of the UKP-SQuARE project (grant GU 798/29-1).

\bibliography{anthology,custom}
\bibliographystyle{acl_natbib}

\onecolumn
\appendix

\section{Performance of Using Different Retrievers for Negative Mining in GPL}
\label{sec:switching_negative_mining_retrievers}
The performance of using different retrievers (BM25, dense and BM25 + dense) for mining hard negatives in GPL is shown in \autoref{tbl:switching_negative_mining_retrievers}. The results show GPL performs best when using hard negatives mined by dense retrievers.

\begin{table*}[ht]
\centering
\resizebox{13.5cm}{!}{
\begin{tabular}{|l|c|c|c|c|c|c|c|} 
\hline
\diagbox{\textbf{Method}}{\textbf{Dataset}} & \textbf{FiQA}          & \textbf{SciFact}       & \textbf{BioASQ} & \textbf{TRECC.}        & \textbf{CQADup.}       & \textbf{Robust04}      & \textbf{Avg.}           \\ 
\hline
GPL (w/ BM25 + dense) & 32.9 & 64.4 & 61.1 & 68.6 & 33.8 & 41.3 & 50.4 \\
GPL (w/ BM25) & 31.1 & 60.9 & 57.8 & 67.5 & 33.5 & 35.9 & 47.8 \\
GPL (w/ dense) & 32.8 & 66.4 & 61.0 & 72.6 & 34.5 & 41.4 & 51.5 \\
\hline
MS MARCO                & 26.7 & 57.1 & 52.9 & 66.1 & 29.6 & 39.0 & 45.2 \\
\hline
\end{tabular}}
\caption{Performance (nDCG@10) of using different retrievers for hard-negative mining in GPL. The scores of the baseline MS MARCO and the scores of GPL with dense retrievers are copied from \autoref{tbl:main_results}.}
\label{tbl:switching_negative_mining_retrievers}
\end{table*}

\section{Performance of the Zero-Shot Retrievers in Hard-Negative Mining}
\label{sec:retrievers_in_gpl}

The performance of directly using the zero-shot retrievers for hard-negative mining in GPL is shown in \autoref{tbl:retrievers_in_gpl}. Compared with the strong baseline (MS MARCO in \autoref{tbl:retrievers_in_gpl}) trained with MarginMSE, \textit{msmarco-distilbert-base-v3} and \textit{msmarco-MiniLM-L-6-v3} are much worse in terms of zero-shot generalization on each dataset. This comparison supports GPL can indeed train powerful domain-adapted dense retrievers with minimum reliance on choices of the retrievers for hard-negative mining.

\begin{table*}[ht]
\centering
\resizebox{13.5cm}{!}{
\begin{tabular}{|l|c|c|c|c|c|c|c|} 
\hline
\diagbox{\textbf{Method}}{\textbf{Dataset}} & \textbf{FiQA}          & \textbf{SciFact}       & \textbf{BioASQ} & \textbf{TRECC.}        & \textbf{CQADup.}       & \textbf{Robust04}      & \textbf{Avg.}           \\ 
\hline
msmarco-distilbert-base-v3 &24.0 &52.3 &45.6 &61.1 &24.3 &30.6 &39.7 \\
msmarco-MiniLM-L-6-v3 &23.3 &48.8 &41.9 &57.9 &24.3 &28.5 &37.5 \\
\hline
MS MARCO                & 26.7 & 57.1 & 52.9 & 66.1 & 29.6 & 39.0 & 45.2 \\
\hline
\end{tabular}}
\caption{Performance (nDCG@10) of different zero-shot retrievers. \textit{msmarco-distilbert-base-v3} and \textit{msmarco-MiniLM-L-6-v3} are used in GPL for hard-negative mining. The scores of the baseline MS MARCO are copied from \autoref{tbl:main_results}.}
\label{tbl:retrievers_in_gpl}
\end{table*}

\section{Target Datasets}
\label{sec:target_datasets}
\textbf{FiQA} is for the task of opinion question answering over financial data. It contains 648 queries and 5.8K passages from StackExchange posts under the Investment topic in the period between 2009 and 2017. The labels are binary (relevant or irrelevant) and there are 2.6 passages in average labeled as relevant for each query.

\textbf{SciFact} is for the task of verifying scientific claims using evidence from the abstracts of the scientific papers. It contains 300 queries and 5.2K passages built from S2ORC~\cite{lo-etal-2020-s2orc}, a publicly-available corpus of millions of scientific articles. The labels are binary and there are 1.1 passages in average labeled as relevant for each query.

\textbf{BioASQ} is for the task of biomedical question answering. It originally contains 500 queries and 15M articles from PubMed\footnote{\url{https://pubmed.ncbi.nlm.nih.gov/}}. The labels are binary and it has 4.7 passages in average labeled as relevant for each query. For efficient training and evaluation, we randomly remove irrelevant passages to make the final corpus size to 1M.

\textbf{TREC-COVID} is an ad-hoc search challenge for scientific articles related to COVID-19 based on the CORD-19 dataset~\cite{wang-etal-2020-cord}. It originally contains 50 queries and 171K documents. The original corpus has many documents with only a title and an empty body. We remove such documents and the final corpus size is 129.2K. The labels in TREC-COVID are 3-level (i.e. 0, 1 and 2) and there are 430.8 passages in average labeled as 1 or 2 in the clean-up version.

\textbf{CQADupStack} is a dataset for community question-answering, built from 12 StackExchange subforums: Android, English, Gaming, Gis, Mathematica, Physics, Programmers, Stats, Tex, Unix, Webmasters and WordPress. The task is to retrieve duplicate question posts with both a title and a body text given a post title. It has 13.1K queries and 457.2k passages. The labels are binary and there are 1.4 passages in average labeled as relevant for each query. As in \citet{thakur2021beir}, the average score of the 12 sub-tasks is reported.

\textbf{Robust04} is a dataset for news retrieval focusing on poorly performing topics. It has 249 queries and 528.2K passages. The labels are 3-level and there are in average 69.9 passages labeled as relevant for each query.

The detailed statistics of these target datasets are shown in \autoref{tbl:statistics}.

\begin{table*}[ht]
\centering
\resizebox{16cm}{!}{
\begin{tabular}{|l|c|c|c|c|c|c|c|c|} 
\hline
\diagbox{\textbf{Dataset}}{\textbf{Statistics}} & \textbf{Domain}          & \textbf{Title}       & \textbf{Relevancy} & \textbf{\#Queries}        & \textbf{\#Passages}       & \textbf{PPQ}      & \textbf{Query Len.} & \textbf{Passage Len.}           \\ 
\hline
FiQA & Financial &\xmark &Binary &648 &57.6K &2.6 &10.8 & 132.2\\
SciFact & Scientific & \cmark &Binary &300 &5.2K &1.1 &12.4 & 213.6 \\
BioASQ & Bio-Medical &\cmark & Binary &500 &1.0M &4.7 &8.1 & 204.1 \\
BioASQ$^*$ & Bio-Medical &\cmark & Binary &500 &14.9M &4.7 &8.1 & 202.6 \\
TREC-COVID & Bio-Medical &\cmark & 3-Level &50 &129.2K &430.8 &10.6 &210.3 \\
TREC-COVID$^*$ & Bio-Medical &\cmark & 3-Level &50 &171.3K &493.5 &10.6 &160.8 \\
CQADupStack &Forum &\cmark &Binary &13,145 &457.2K &1.4 &8.6 &129.1 \\
Robust04 &News &\xmark &3-Level &249 &528.2K &69.9 &15.3 &466.4 \\
\hline
\end{tabular}}
\caption{Statistics of the target datasets used in the experiments. Column \textbf{Title} indicates whether there is (\cmark) a title for each passage or not (\xmark). Column \textbf{PPQ} represents number of Passages Per Query. Query/passage lengths are counted in words. Symbol $*$ marks the original version from the BeIR benchmark~\cite{thakur2021beir}}
\label{tbl:statistics}
\end{table*}

We also evaluate the models trained in this work on the original version of BioASQ and TREC-COVID datasets from BeIR~\cite{thakur2021beir}. The results are shown in \autoref{tbl:full_beir}.



\section{Results on full BeIR}
\label{sec:full_beir}
We also evaluate the models on all the 18 BeIR datasets. We include DocT5Query~\cite{nogueira2019doc2query}, the strong baseline based on document expansion with the T5 query generator (also used in GPL for query generation) + BM25 (Anserini). We also include the powerful zero-shot model \textbf{TAS-B}~\cite{DBLP:conf/sigir/HofstatterLYLH21}, which is trained on MS MARCO with advanced knowledge-distillation techniques into comparison. Viewing TAS-B as the base model and also the negative miner, we apply QGen and GPL on top of them, resulting in \textbf{TAS-B + QGen} and \textbf{TAS-B + GPL}, resp.

The results are shown in~\autoref{tbl:full_beir}. We find both DocT5Query and BM25 (Anserini) outperform MS MARCO, TSDAE and QGen, in terms of both average performance and average rank. QGen struggles to beat MS MARCO, the zero-shot baseline and it even significantly harms the performance on many datasets, e.g. TREC-COVID, FEVER, HotpotQA, NQ. \citet{thakur2021beir} also observes the same issue, claiming that the bad generation quality on these corpora is the key to the failure of QGen. On the other hand, GPL significantly outperforms these baselines above, achieving average rank 5.2 and can consistently improve the performance over the zero-shot model on all the datasets. For TSDAE, TSDAE + QGen and TSDAE + GPL, the conclusion remains the same as in the main paper.

For the powerful zero-shot model TAS-B, it outperforms QGen and performs on par with TSDAE + QGen. When building on top of TAS-B, GPL can also yield significant performance gain by up-to 21.5 nDCG@10 points (on TREC-COVID) and 4.6 nDCG@10 points on average. This TAS-B + GPL model performs the best over all these retriever models, achieving the averaged rank equal to 3.2. However, when applying QGen on top of TAS-B, it cannot improve the overall performance but also harms the individual performance on many datasets, instead.

\begin{table}
\centering
\resizebox{16cm}{!}{
\begin{tabular}{|l|c|c|c|c|c|c|c|c|c|c|c||c|} 
\hline
\diagbox{\textbf{Dataset}}{\textbf{Method}}  & \begin{tabular}[c]{@{}c@{}}BM25 \\(Anserini)\end{tabular} & \begin{tabular}[c]{@{}c@{}}DocT5-\\Query\end{tabular} & \begin{tabular}[c]{@{}c@{}}MS \\MARCO\end{tabular} & TSDAE         & QGen & \begin{tabular}[c]{@{}c@{}}TSDAE + \\QGen (Ours)\end{tabular} & GPL           & \begin{tabular}[c]{@{}c@{}}TSDAE + \\GPL\end{tabular} & TAS-B         & \begin{tabular}[c]{@{}c@{}}TAS-B + \\QGen\end{tabular} & \begin{tabular}[c]{@{}c@{}}TAS-B + \\GPL\end{tabular} & \begin{tabular}[c]{@{}c@{}}BM25 + CE\\(Upperbound)\end{tabular}  \\ 
\hline
\textbf{FiQA}       & 23.6                                                      & 29.1                                                  & 26.7                                               & 29.3          & 28.7 & 31.4                                                          & 32.8          & \textbf{34.4}                                         & 29.8          & 30.1                                                   & \textbf{34.4}                                         & 34.7                                                             \\
\textbf{SciFact}    & 66.5                                                      & 67.5                                                  & 57.1                                               & 62.8          & 63.8 & 66.7                                                          & 66.4          & \textbf{68.9}                                         & 63.5          & 65.3                                                   & 67.4                                                  & 68.8                                                             \\
\textbf{BioASQ$^*$} & \textbf{46.5}                                             & 43.1                                                  & 33.6                                               & 37.3          & 36.9 & 38.5                                                          & 41.2          & 40.9                                                  & 36.2          & 38.5                                                   & 44.2                                                  & 52.3                                                             \\
\textbf{TRECC.$^*$} & 65.6                                                      & 71.3                                                  & 66.1                                               & 70.8          & 56.0 & 58.4                                                          & 71.8          & \textbf{74.9}                                         & 48.5          & 56.6                                                   & 70.0                                                  & 75.7                                                             \\
\textbf{CQADup.}    & 29.9                                                      & 32.5                                                  & 29.6                                               & 31.8          & 33.0 & 35.3                                                          & 34.5          & 35.1                                                  & 31.5          & 33.7                                                   & \textbf{35.7}                                         & 37.0                                                               \\
\textbf{Robust04}   & 40.8                                                      & \textbf{43.7}                                         & 39.0                                               & 39.4          & 38.1 & 37.4                                                          & 41.4          & 43.0                                                    & 42.4          & 39.4                                                   & \textbf{43.7}                                         & 47.5                                                             \\
\textbf{ArguAna}    & 41.4$^\dagger$                                                      & 46.9$^\dagger$                                                  & 33.9                                               & 37.5          & 52.4 & 54.7                                                          & 48.3          & 51.2                                                  & 43.4          & 51.8                                                   & \textbf{55.7}                                         & 41.7$^\dagger$                                                             \\
\textbf{Climate-F.} & 21.3                                                      & 20.1                                                  & 20.0                                               & 16.8          & 22.5 & 22.6                                                          & 22.7          & 22.2                                                  & 22.1          & \textbf{24.4}                                          & 23.5                                                  & 25.3                                                             \\
\textbf{DBPedia}    & 31.3                                                      & 33.1                                                  & 34.2                                               & 35.4          & 33.1 & 33.2                                                          & 36.1          & 36.1                                                  & \textbf{38.4} & 32.7                                                   & \textbf{38.4}                                         & 40.9                                                             \\
\textbf{FEVER}      & 75.3                                                      & 71.4                                                  & 76.5                                               & 64.0          & 63.8 & 64.2                                                          & \textbf{77.9} & 78.6                                                  & 69.5          & 63.9                                                   & 75.9                                                  & 81.9                                                             \\
\textbf{HotpotQA}   & 60.3                                                      & 58.0                                                  & 55.4                                               & \textbf{63.8} & 51.4 & 52.2                                                          & 56.5          & 57.2                                                  & 58.4          & 52.0                                                   & 58.2                                                  & 70.7                                                             \\
\textbf{NFCorpus}   & 32.5                                                      & 32.8                                                  & 27.7                                               & 31.2          & 31.4 & 33.7                                                          & 34.2          & 33.9                                                  & 31.9          & 33.4                                                   & \textbf{34.5}                                         & 35.0                                                               \\
\textbf{NQ}         & 32.9                                                      & 39.9                                                  & 45.6                                               & 47.1          & 35.4 & 34.6                                                          & 46.7          & 47.1                                                  & 46.3          & 36.3                                                   & \textbf{48.3}                                         & 53.3                                                             \\
\textbf{Quora}      & 78.9                                                      & 80.2                                                  & 81.2                                               & 83.3          & 85.0 & \textbf{85.7}                                                 & 83.2          & 83.1                                                  & 83.5          & 85.3                                                   & 83.6                                                  & 82.5                                                             \\
\textbf{SciDocs}    & 15.8                                                      & 16.2                                                  & 13.6                                               & 15.4          & 15.5 & \textbf{17.1}                                                 & 16.1          & 16.8                                                  & 14.9          & 16.4                                                   & 16.9                                                  & 16.6                                                             \\
\textbf{Signal-1M}  & \textbf{33.0}                                             & 30.7                                                  & 24.4                                               & 25.9          & 26.8 & 26.8                                                          & 26.5          & 27.6                                                  & 28.9          & 26.6                                                   & 27.6                                                  & 33.8                                                             \\
\textbf{TRECN.}     & 39.8                                                      & 42.0                                                  & 36.0                                               & 35.0          & 36.0 & 38.3                                                          & 40.7          & 41.5                                                  & 37.7          & 38.0                                                   & \textbf{42.1}                                         & 43.1                                                             \\
\textbf{Touché20}   & \textbf{36.7}                                             & 34.7                                                  & 19.6                                               & 21.8          & 17.1 & 17.2                                                          & 23.1          & 23.5                                                  & 16.2          & 17.5                                                   & 25.5                                                  & 27.1                                                             \\ 
\hline
\textbf{Avg.}       & 42.9                                                      & 44.1                                                  & 40.0                                               & 41.6          & 40.4 & 41.6                                                          & 44.5          & 45.3                                                  & 41.3          & 41.2                                                   & \textbf{45.9}                                         & 48.2                                                             \\
\textbf{Avg. Rank}  & 7.6                                                       & 6.2                                                   & 9.8                                                & 8.2           & 8.9  & 6.5                                                           & 5.2           & 4.2                                                   & 7.8           & 7.3                                                    & \textbf{3.2}                                          & 2.4                                                              \\
\hline
\end{tabular}}
\caption{Performance (nDCG@10) on all the original 18 BeIR datasets. The results of MS MARCO, TSDAE, QGen, TSDAE + QGen, GPL and TSDAE + GPL on FiQA, SciFact, CQADupStack and Robust04 are copied from~\autoref{tbl:main_results}. The results of BM25, DocT5Query and BM25 + CE come from~\citet{thakur2021beir}. $\dagger$ marks correction over the original scores, where identical IDs between queries and passages are removed. TRECN. is short for TREC-NEWS. Avg. Rank is the average over the rank of the performance on each dataset over the different models (the lower, the better).}
\label{tbl:full_beir}
\end{table}

\section{Performance of Unsupervised Pre-Training}
\label{sec:unsupervised}

The performance of the unsupervised pre-training methods without access to the MS MARCO data is shown in \autoref{tbl:unsupervised}. We find ICT is the best method, achieving highest scores on all the datasets. However, all the unsupervised pre-training methods cannot directly yield improvement in performance compared with the zero-shot baseline. 

\begin{table*}[ht]
\centering
\resizebox{13cm}{!}{
\begin{tabular}{|l|c|c|c|c|c|c|c|} 
\hline
\diagbox{\textbf{Method}}{\textbf{Dataset}} & \textbf{FiQA}          & \textbf{SciFact}       & \textbf{BioASQ} & \textbf{TRECC.}        & \textbf{CQADup.}       & \textbf{Robust04}      & \textbf{Avg.}           \\ 
\hline
CD &6.6 &0.6 &0.3 &9.8 &8.1 &3.8 &4.9 \\
CT &0.2 &0.7 &0.0 &2.5 &0.9 &0.0 &0.7 \\
MLM &5.4 &27.8 &4.7 &16.0 &8.5 &6.1 &11.4 \\
TSDAE &7.8 &37.2 &6.9 &9.4 &14.3 &10.1 &14.3 \\
SimCSE &5.5 &25.0 &13.1 &26.0 &14.6 &9.8 &15.7 \\
ICT &10.2 &42.6 &39.0 &47.5 &23.0 &16.5 &29.8 \\
\hline
MS MARCO                & 26.7 & 57.1 & 52.9 & 66.1 & 29.6 & 39.0 & 45.2 \\
\hline
\end{tabular}}
\caption{Performance (nDCG@10) of unsupervised pre-training methods with only access to the target corpus as the training data. The scores of the zero-shot baseline MS MARCO are copied from \autoref{tbl:main_results}.}
\label{tbl:unsupervised}
\end{table*}

\section{Examples of Generated Queries under Different Temperatures}
\label{sec:temp_examples}
The generation temperature controls the sharpness of the next-token distribution. The examples for one passage from FiQA are shown in \autoref{tbl:temp_examples} Higher temperature results in longer and less duplicate queries under more risk of generating non-sense texts.

\begin{table*}[t]
\centering
\resizebox{14.5cm}{!}{
\begin{tabular}{|l|c|c|} 
\hline
\textbf{Item}             & \textbf{Text}  & \textbf{Pseudo Label}                                                                                                                                                                                                                                                                                                                                                                                                                                                                    \\ 
\hline
\textbf{Input Passage }  & \begin{tabular}[c]{@{}c@{}}You can never use a health FSA for individual health insurance premiums. Moreover,\\ FSA plan sponsors can limit what they are will to reimburse. While you can't use a health\\ FSA for premiums, you could previously use a 125 cafeteria plan to pay premiums, but it\\ had to be a separate election from the health FSA. However, under N. 2013-54, even\\ using a cafeteria plan to pay for indivdiual premiums is effectively prohibited.\end{tabular}  & -- \\ 
\hline
\multirow{3}{*}{\textbf{Temperature 0.1}}  & can you use a cafeteria plan for premiums                                                                                               & 9.1                                                                                                                                                                                                                                                                                                                                                  \\ 
\cline{2-3}
                          & can you use a cafeteria plan for premiums                                                                              & 9.1                                                                                                                                                                                                                                                                                                                                                                   \\ 
\cline{2-3}
                          & can you use a cafeteria plan for premiums                                                                            & 9.1                                                                                                                                                                                                                                                                                                                                                                     \\ 
\hline
\multirow{3}{*}{\textbf{Temperature 1.0}}  & can i use my fsa to pay for a health plan                                                                                         & 9.7                                                                                                                                                                                                                                                                                                                                                        \\ 
\cline{2-3}
                          & can i use my health fsa for an individual health plan?                                                                   & 9.9                                                                                                                                                                                                                                                                                                                                                                 \\ 
\cline{2-3}
                          & can fsa pay premiums     
                          & 9.2\\ 
\hline
\multirow{5}{*}{\textbf{Temperature 3.0}}  & cafe a number cafe plan is used by                                                                                 & -10.5                                                                                                                                                                                                                                                                                                                                                                       \\ 
\cline{2-3}
                          & \begin{tabular}[c]{@{}c@{}}what type of benefits do the health savings accounts cover \\when applying for medical terms health insurance\end{tabular}                                                       & -7.2                                                                                                                                                                                                                                                                              \\ 
\cline{2-3}
                          & \begin{tabular}[c]{@{}c@{}}why can't an individual file medical premium on their insurance account with an fsa plan\\ instead of healthcare policy.\end{tabular}                                        & 6.0                                                                                                                                                                                                                                                                                  \\ 
\hline
\multirow{3}{*}{\textbf{Temperature 5.0}}  & which one does not apply after an emergency medical                                                                         & -11.1                                                                                                                                                                                                                                                                                                                                                              \\ 
\cline{2-3}
                          & is medicare cafe used exclusively as plan funds (health savings account                                                 & -7.2                                                                                                                                                                                                                                                                                                                                                                  \\ 
\cline{2-3}
                          & how soon to transfer coffee bean fses to healthcare                                                                  & -11.0                                                                                                                                                                                                                                                                                                                                                                     \\ 
\hline
\multirow{5}{*}{\textbf{Temperature 10.0}} & \begin{tabular}[c]{@{}c@{}}will employer limit premiums reimbursement on healthcare expenses with caeatla\\ cafetaril and capetarians account on my employer ca. plans and deductible accounts\\ a.f,haaq and asfrhnta,\end{tabular}                                                                                                                                     & -2.5                                                                                                                 \\ 
\cline{2-3}
                          & kfi what is allowed as personal health account or ca                                                                        & -10.2                                                                                                                                                                                                                                                                                                                                                              \\ 
\cline{2-3}
                          & \begin{tabular}[c]{@{}c@{}}do people put funds back to buy plan plans before claiming an deductible without the\\ provider or insurance cover f/f associator funds of the person you elect? healthfin depto\\ of benefit benefits deduct all oe premiumto payer for individual care\end{tabular}                                                                                                   & -4.5                                                                                       \\
\hline
\end{tabular}}
\caption{Examples of generated queries under different temperature value for a passage from FiQA.}
\label{tbl:temp_examples}
\end{table*}

\end{document}